\documentclass[11pt]{article}

\usepackage[final]{acl}

\usepackage{times}
\usepackage{latexsym}

\usepackage[T1]{fontenc}

\usepackage[utf8]{inputenc}

\usepackage{microtype}

\usepackage{graphicx}

\usepackage{float}
\usepackage{enumitem}

\usepackage{amsmath}
\usepackage{amssymb}
\usepackage{booktabs}
\usepackage{multirow}
\usepackage{url}
\usepackage{algorithm}
\usepackage{algpseudocode}

\newcommand\blfootnote[1]{%
  \begingroup
  \renewcommand\thefootnote{}\footnote{#1}%
  \addtocounter{footnote}{-1}%
  \endgroup
}

\title{FormulaSPIN: Self-Play Fine-Tuning for Natural Language to Spreadsheet Formula Generation}

\author{Cy Xie \\
  Cupertino, CA \\
  \texttt{ctrom0604@gmail.com}}

\begin{document}
\maketitle

\begin{abstract}
Spreadsheet applications are used by hundreds of millions worldwide, yet writing formulas remains a significant barrier. Existing approaches rely on static supervised data, which quickly saturates on limited annotations. In this paper, we introduce \textsc{FormulaSPIN}, a self-play framework that breaks the ceiling of supervised fine-tuning by enabling iterative self-improvement without any additional data. Vanilla SPIN fails on this task: it uniformly penalizes every non-matching output, so execution-equivalent alternatives are pushed down as negatives in one example while serving as ground truth in another, producing contradictory gradients. Our framework resolves this by exploiting formula generation's unique advantage: binary executability provides implicit supervision that separates semantic errors from valid stylistic variants. We frame training as a two-player game in which the main player learns to prefer ground-truth formulas over those from its previous version, while execution feedback sorts outputs into distinct granularities---enabling an adaptive curriculum that shifts from semantic correctness to stylistic refinement. To further increase accuracy, we incorporate ExecVote, a semantic-level voting mechanism that naturally handles multiple valid formulations. Experiments on multiple benchmarks demonstrate that \textsc{FormulaSPIN} achieves state-of-the-art performance, with 74.9\% exact match and 87.1\% execution accuracy on NL2FORMULA, matching models trained with additional preference annotations while outperforming both traditional SFT and frontier proprietary models. These findings underscore self-play's potential to tackle scarce data tasks and open the door to extending it beyond executable domains.
\end{abstract}

\blfootnote{Code Implementation is available at \url{https://github.com/xTronzZ/FormulaSPIN}.}

\section{Introduction}

Spreadsheet applications like Microsoft Excel and Google Sheets are ubiquitous tools for data analysis, used by hundreds of millions worldwide. However, writing formulas remains a significant barrier for many users, particularly when dealing with complex operations involving multiple functions, conditional logic, and cell references \citep{Gulwani2011}. Recent work addresses this via Natural Language to Formula (NL2Formula) generation—first systematically benchmarked by \citet{Zhao2024nl2formula}—where users express their intent in plain language and the system automatically produces executable formulas.

\begin{figure}[t]
\centering
\includegraphics[width=\columnwidth]{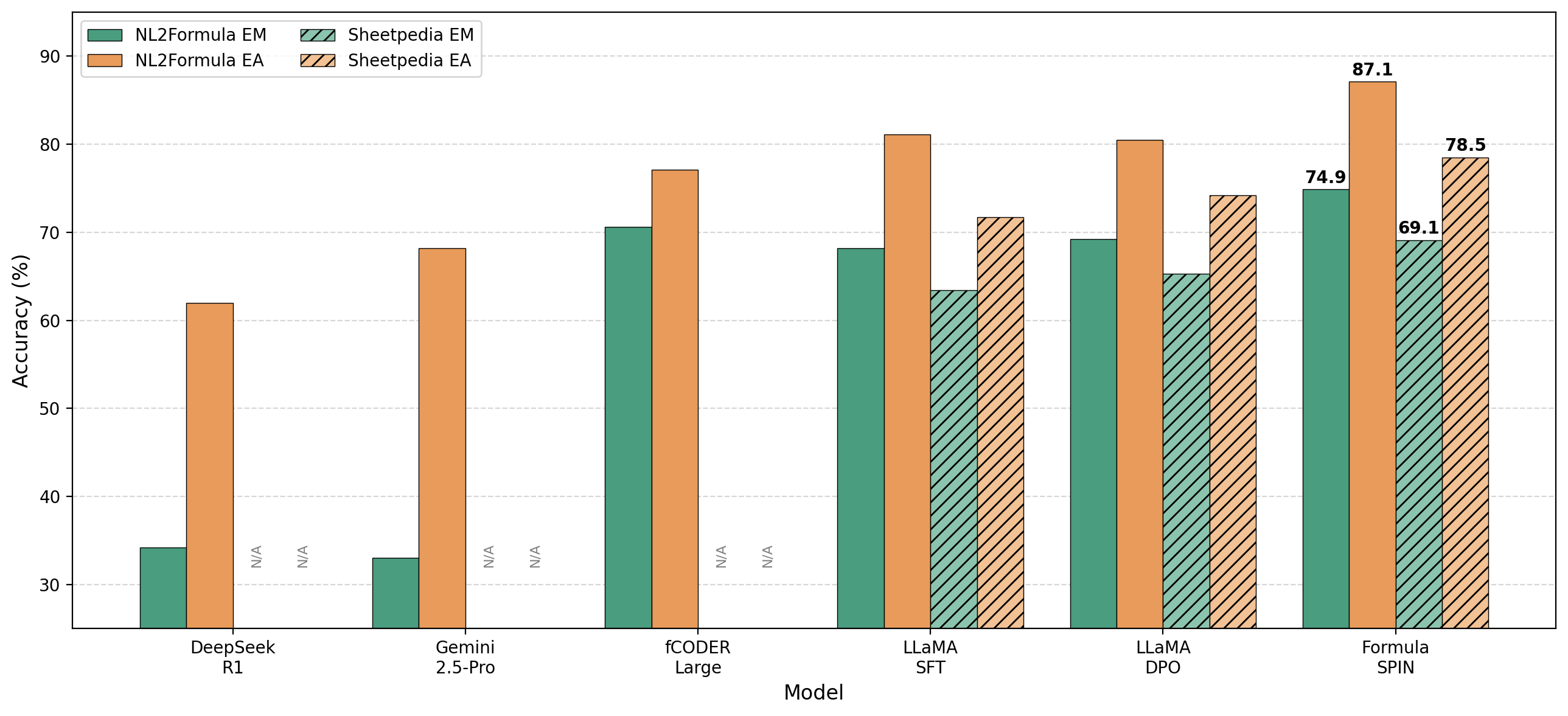}
\caption{Performance comparison on NL2Formula-70K~\citep{Zhao2024nl2formula} and Sheetpedia~\citep{Tian2025sheetpedia}. FormulaSPIN achieves state-of-the-art results on both benchmarks without additional annotations, outperforming the SFT baseline by +6.7\% EM (Exact Match) / +6.0\% EA (Execution Accuracy) on NL2Formula-70K and +5.7\% EM / +6.8\% EA on Sheetpedia, while matching SFT-DPO trained with GPT-4.1-labeled preference pairs as well as strong proprietary models such as Gemini-2.5-Pro and DeepSeek-R1.}
\label{fig:benchmark_combined}
\end{figure}

Despite promising progress, most NL2Formula systems rely on supervised fine-tuning on static datasets, framing formula generation as a conventional seq2seq problem. However, existing datasets are limited in scale and coverage: the largest benchmark contains only 70K examples spanning limited formula patterns and table structures. Although recent work Sheetpedia \citep{Tian2025sheetpedia} provides over 290K worksheets, it serves as a general-purpose resource; fine-tuning on it yields only 2.1\% improvement over the 70K benchmark despite the much larger corpus. This indicates that simply scaling static data offers limited benefit, as training on fixed datasets quickly plateaus, with additional epochs producing diminishing returns or even degrading performance \citep{Chen2024spin}. Static supervision offers no clear path forward: collecting preference data for RLHF or RLAIF reintroduces the same bottleneck, requiring extensive manual evaluation or access to proprietary models. Current systems thus remain far below their potential, particularly struggling with complex nested formulas and ambiguous queries where supervision is scarcest. Overcoming this ceiling requires a fundamentally different paradigm, enabling models to improve beyond initial supervision without additional annotations or external preference signals.

To address these challenges, we introduce FormulaSPIN, a self-play fine-tuning framework specifically designed for spreadsheet formula generation. Drawing inspiration from AlphaGo Zero's self-play mechanism \citep{Silver2017} and recent advances in self-improving language models \citep{Chen2024spin}, our approach enables iterative model improvement through a two-player game: the current model (main player) learns to distinguish formulas generated by its previous version (opponent) from ground-truth formulas, while the opponent strives to generate formulas indistinguishable from human-written ones.

The key insight underlying FormulaSPIN is that formula generation offers a unique advantage over general generation tasks: executability provides implicit supervision. Unlike open-ended text where quality assessment requires human judgment, formulas can be automatically validated by executing them in a spreadsheet engine. When compared to code generation that also incorporates execution-based supervision, formula execution is \emph{binary, deterministic, and lightweight}, requiring no manually designed test cases or complex runtime environments. Critically, vanilla self-play fails for this task because it treats all non-matching formulas uniformly, penalizing execution-equivalent alternatives (e.g., \texttt{SUM(A1:A5)} vs. \texttt{A1+A2+A3+A4+A5}) and creating contradictory training signals. 

We address this through three novel components:
\begin{itemize}
    \item \textbf{Formula-Aware Self Play}: incorporates execution feedback into the self-play objective, categorizing generated formulas by error type to weight their contribution appropriately.
    \item \textbf{Multi-Granularity Curriculum}: automatically adjusts training focus based on the distribution of semantic errors versus stylistic variations, progressing from correctness to style as the model matures.
    \item \textbf{Execution-based Voting}: generates multiple candidates at inference time and selects the best one via execution-based semantic voting.
\end{itemize}

Our contributions are fourfold. 
\textbf{(1)} We identify a fundamental limitation of supervised learning for NL2Formula and reformulate spreadsheet formula generation as an executable self-improvement problem, where static annotations alone are insufficient for learning complex formulas and ambiguous queries. 
\textbf{(2)} We propose FormulaSPIN, a self-play fine-tuning framework that leverages the binary executability of formulas to provide intrinsic supervision, enabling iterative improvement without additional human preferences or external teacher models. 
\textbf{(3)} We introduce an adaptive semantic-to-stylistic curriculum that automatically shifts training focus from correctness to canonical formulation as the model improves. 
\textbf{(4)} Through extensive in-domain and out-of-domain experiments, we demonstrate that FormulaSPIN substantially outperforms supervised fine-tuning and matches preference-optimized models trained with large-scale additional annotations, with particularly strong gains on complex nested formulas.

\section{Related Work}

Our work addresses three fundamental challenges in formula generation: \textit{(i)} limited training data that constrains model performance, \textit{(ii)} difficulty in obtaining quality supervision signals, and \textit{(iii)} handling multiple valid solutions at inference time. We organize related work around how prior approaches tackle these challenges and where they fall short.

\subsection{Formula Generation and Data Bottleneck}

Semantic parsing evolved from logic forms \citep{Price1990,Zelle1996} through knowledge-base grounding \citep{Berant2014} to SQL-based systems \citep{Zhong2017,Yu2018}. For formula synthesis, FlashFill \citep{Gulwani2011} pioneered programming-by-example. Structure-aware pre-training methods model table layouts \citep{Wang2021tuta,Liu2022tapex} and formula semantics \citep{Cheng2022fortap}. Recently, \citet{Zhao2024nl2formula} introduced NL2Formula with 70K examples, and \citet{Tian2025sheetpedia} constructed Sheetpedia with 290K samples.

\begin{figure*}[h]
\centering
\includegraphics[width=1\textwidth]{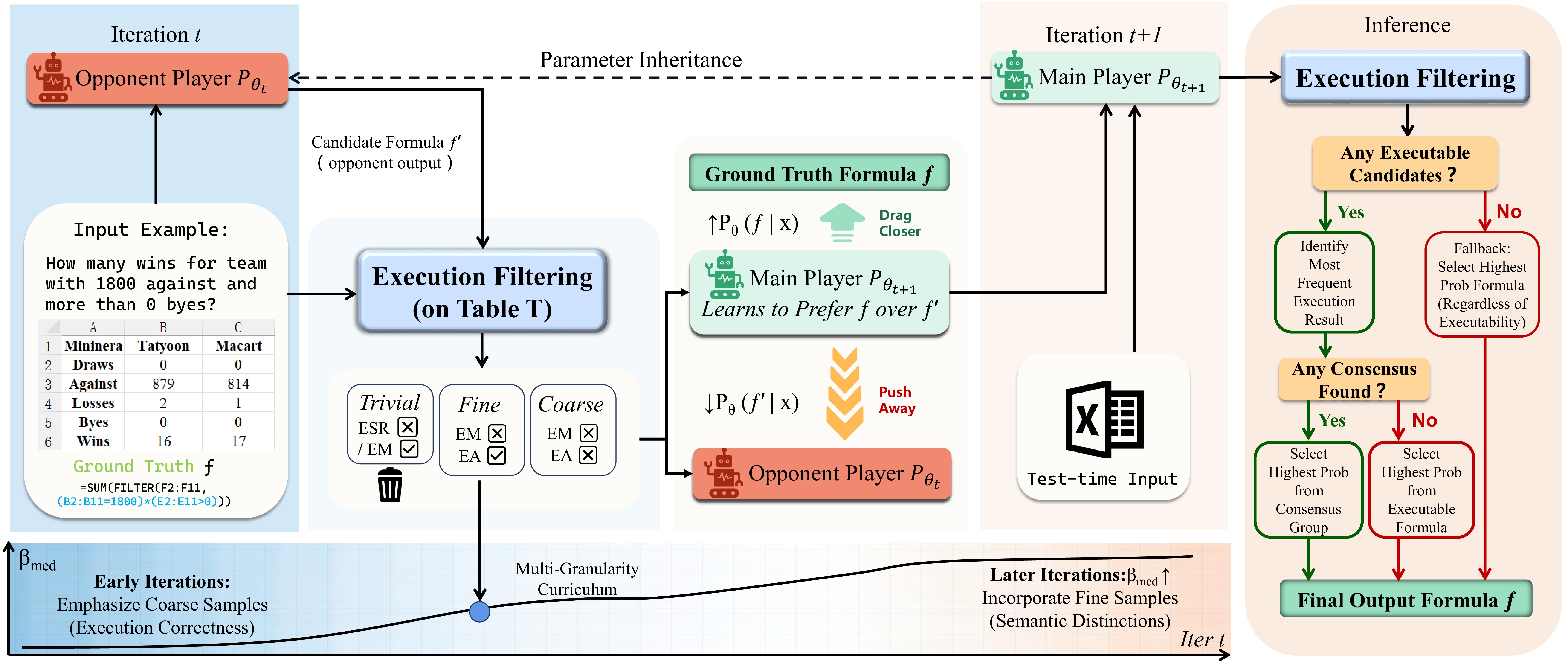}
\caption{\textbf{The FormulaSPIN framework.} The model iteratively self-improves via a two-player game, utilizing Execution Filtering to construct an adaptive curriculum that progresses from semantic correctness (\textit{Coarse} samples) to stylistic refinement (\textit{Fine} samples). At inference, an execution-based voting strategy selects the final formula based on execution agreement among sampled candidates.}
\label{fig:pipeline}
\end{figure*}

These approaches face inevitable saturation on fixed datasets, particularly for complex nested formulas where supervision is scarcest. Our work breaks this ceiling through iterative self-improvement, generating training signal from the model's own outputs without additional annotation.

\subsection{From Supervision to Self-Improvement}
Given the data bottleneck, recent work explores alternative supervision: external preference collection  versus self-generated feedback.

Preference-based methods \citep{Ouyang2022,Bai2022constitutional,Rafailov2023dpo} require expensive human annotations or proprietary teacher models, shifting the bottleneck. For formula generation, distilling GPT-4.1 introduces external dependencies and annotation costs.

Self-play mechanisms, from TD-Gammon \citep{Tesauro1995} to AlphaGo \citep{Silver2017}, enable improvement by competing against oneself. SPIN \citep{Chen2024spin} adapted this to language models, but vanilla SPIN fails for formula generation: it treats non-matching outputs uniformly, penalizing execution-equivalent alternatives (e.g., \texttt{SUM(A1:A5)} vs. \texttt{A1+A2+A3+A4+A5}) and creating contradictory gradients.

We introduce formula-aware self-play that exploits execution feedback to distinguish semantic errors from stylistic variations, enabling stable learning without external supervision.

\subsection{Execution-Based Learning for Formulas}

Execution-based validation has proven effective in code generation \citep{le2022coderl,Chen2024selfdebug}, but code execution requires test suites and runtime environments. For spreadsheet formulas, recent work has explored domain-specific reinforcement learning with execution feedback. Fortune \citep{Cao2025fortune} applies PPO with symbol-level rewards that encourage function-argument compatibility and penalize syntax errors. However, such fine-grained reward shaping demands extensive domain knowledge and remains sensitive to reward design choices. Prior spreadsheet work either ignores executability or uses it only for post-hoc filtering \citep{Cheng2022fortap,Liu2022tableformer}.

Formula execution is binary, deterministic, and lightweight—a single table provides complete validation. We exploit this to filter and categorize formulas into semantic versus stylistic errors, constructing an adaptive curriculum. Unlike Fortune's symbol-level rewards or general RL methods \citep{Shojaee2023execution,Shinn2024reflexion} that use sparse binary rewards, our preference-based objective provides richer learning signals by contrasting model outputs with ground-truth references. This also contrasts with constrained decoding \citep{Scholak2021picard} that guarantees syntax but not semantics.

\subsection{Test-Time Scaling by Semantic Consensus}

Inference-time computation improves generation quality \citep{Snell2024scaling}, but existing strategies have limitations. Best-of-N sampling \citep{Cobbe2021} requires trained verifiers; self-consistency \citep{Wang2023selfconsistency} votes on surface forms, missing semantically equivalent solutions. For code, sampling multiple solutions boosts performance \citep{Chen2021codex}, but verification demands comprehensive test suites.

We introduce ExecVote (Execution-based Voting) that votes over execution results rather than surface forms. By grouping formulas into semantic equivalence classes through execution, we naturally handle multiple valid formulations without trained verifiers. Among execution-equivalent candidates, we select the highest-probability formula, combining semantic correctness with stylistic preference.

\section{FormulaSPIN}

\subsection{Problem Formulation}

Given a spreadsheet table $\mathcal{T}$ and a natural language query $q$, the NL2Formula task aims to generate an executable formula $f$ that fulfills the user intent. We denote the training dataset as $\mathcal{D} = \{(q_i, \mathcal{T}_i, f_i)\}_{i=1}^N$, where the model learns a conditional distribution $p_\theta(f \mid q, \mathcal{T})$. Each table is serialized in row-major order with cell coordinates (e.g., ``A1:Revenue | B1:2023 | \ldots'').

\subsection{Formula-Aware Self Play}
We adopt a self-play fine-tuning framework inspired by SPIN~\citep{Chen2024spin}, tailored to spreadsheet formula generation. At iteration $t$, the old model $p_{\theta_t}$ acts as opponent player and generates candidate formulas $f'$ for a given query $q$ and table $\mathcal{T}$, while new model $p_{\theta_{t+1}}$ serves as the main player and is trained to assign higher probability to reference formula $f$ than to $f'$. As self-play progresses, the opponent produces increasingly challenging formulas, pushing the main player toward finer-grained semantic understanding.

\begin{figure}[t]
\centering
\includegraphics[width=\columnwidth]{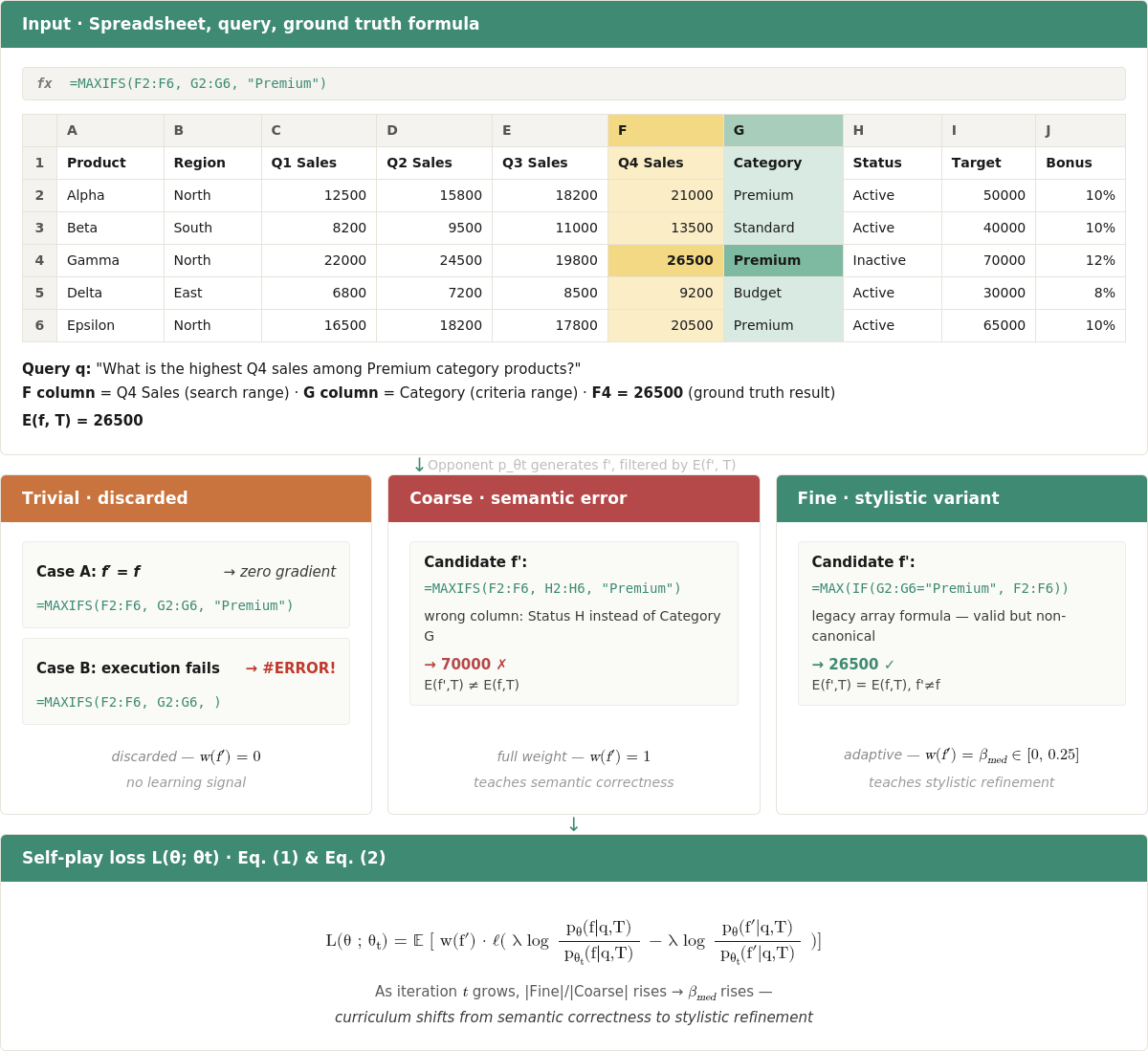}
\caption{\textbf{Execution filtering and the Formula-Aware Self-Play objective.} 
The opponent $p_{\theta_t}$ generates $f'$, which is sorted by executing against $\mathcal{T}$: \emph{Trivial} samples (exact matches or execution failures) are discarded; \emph{Coarse} samples (wrong result) receive full weight to teach semantic correctness; \emph{Fine} samples (correct result, non-canonical form) receive adaptive weight $\beta_{\text{med}}$ for stylistic refinement.}
\label{fig:execution_filtering}
\end{figure}

A key advantage of formula generation is deterministic executability, which we leverage through \emph{execution filtering}: comparing $\mathcal{E}(f', \mathcal{T})$ with $\mathcal{E}(f, \mathcal{T})$ sorts each candidate into \emph{Trivial}, \emph{Coarse}, or \emph{Fine} samples (Figure~\ref{fig:execution_filtering}). Execution failures are folded into Trivial because their diverse error patterns provide inconsistent gradients, and the SFT model already achieves $\sim$96\% execution success. The Coarse/Fine split lets us disentangle semantic errors from stylistic variations and weight them differently in the loss.

Building on this filtering mechanism, we measure the relative
preference between models using the log-probability difference
\begin{equation}
\resizebox{\linewidth}{!}{$
\displaystyle
\mathcal{L}(\theta;\, \theta_t)
= \mathbb{E}\!\left[
  w(f') \cdot \ell\!\left(
    \lambda \log \frac{p_{\theta}(f \mid q, \mathcal{T})}
                      {p_{\theta_t}(f \mid q, \mathcal{T})}
    - \lambda \log \frac{p_{\theta}(f' \mid q, \mathcal{T})}
                        {p_{\theta_t}(f' \mid q, \mathcal{T})}
  \right)
\right],
$}
\end{equation}
where the expectation is computed over the distribution
$q \sim \mathcal{Q}(\cdot)$,
$f \sim p_{\text{data}}(\cdot \mid q, \mathcal{T})$,
$f' \sim p_{\theta_t}(\cdot \mid q, \mathcal{T})$,
and $\ell(t) := \log\bigl(1 + \exp(-t)\bigr)$ denotes the
logistic loss.

The relative form $r(f) - r(f')$ encodes the
adversarial dynamic: the main player must outperform the
opponent's ability to distinguish reference from synthetic
formulas, and as the opponent strengthens, the margin for
improvement narrows, compelling progressively finer
discrimination. The weight $w(f')$ is determined by execution
filtering:
\begin{equation}
{\small
\setlength{\arraycolsep}{2pt}
\renewcommand{\arraystretch}{0.85}
w(f') =
\begin{cases}
  1,                  & \text{if } f' \text{ is \emph{Coarse}}, \\
  \beta_{\text{med}}, & \text{if } f' \text{ is \emph{Fine}}, \\
  0,                  & \text{if } f' \text{ is \emph{Trivial}}.
\end{cases}
}
\end{equation}
i.e., Coarse formulas receive full weight, Trivial formulas
receive zero weight, and Fine formulas receive an adjustable
weight $\beta_{\text{med}} \in [0, 0.25]$.

\begin{figure}[t]
  \centering
  \includegraphics[width=\linewidth]{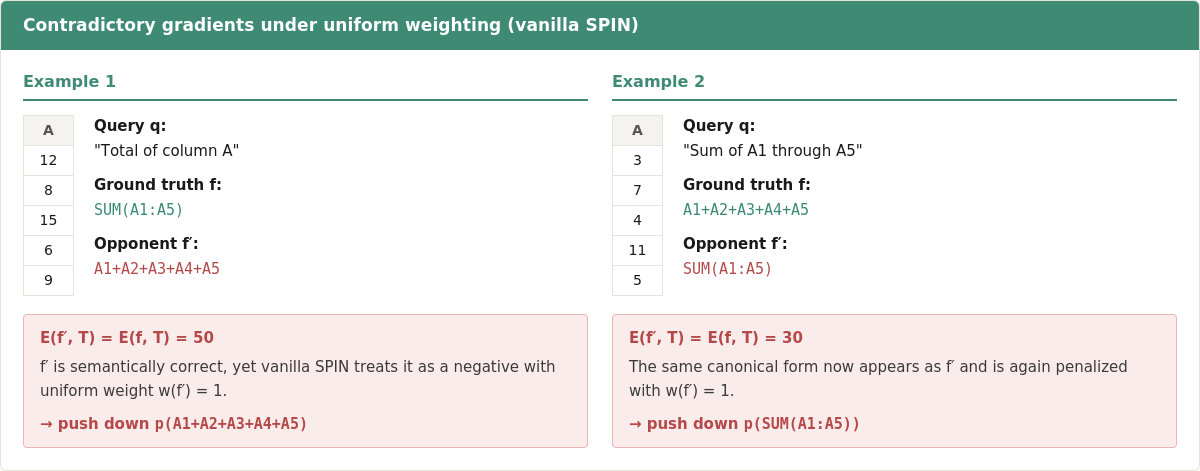}
  \caption{Contradictory gradients under uniform weighting. When \texttt{SUM(A1:A5)} and \texttt{A1+A2+A3+A4+A5} swap roles across examples, vanilla SPIN pushes down both correct forms, destabilizing training.}
  \label{fig:contradictory_gradients}
\end{figure}

The choice of $\beta_{\text{med}}$ for Fine samples addresses a key tension: While Fine samples are execution-equivalent to the reference and thus semantically correct, the self-play objective still treats them as negatives to be pushed away. Assigning them the same weight as Coarse samples would create contradictory gradients---the same canonical form may appear as ground truth in one example and as a penalized opponent output in another (see Figure~\ref{fig:contradictory_gradients})---destabilizing training. Yet setting their weight to zero is also undesirable: reference formulas are typically more concise and idiomatic than execution-equivalent alternatives (e.g., \texttt{SUM(A1:A5)} vs.\ \texttt{A1+A2+A3+A4+A5}), and we want the model's outputs to converge toward these canonical forms rather than drift into verbose variants. 

We resolve this tension by down-weighting Fine samples with $\beta_{\mathrm{med}} \in [0, 0.25]$, and letting $\beta_{\mathrm{med}}$ grow with the Fine/Coarse ratio: early on, when Coarse dominates, we avoid penalizing correct alternatives almost entirely; later, once semantic errors are rare, a mild penalty on non-canonical forms nudges the model toward idiomatic style without conflicting with the correctness objective.
We formalize this as:
\begin{equation}
{\small
\beta_{\text{med}}^{(t)} = \beta_{\max} \cdot
\frac{|\mathcal{S}_{\text{fine}}^{(t)}|}
     {|\mathcal{S}_{\text{fine}}^{(t)}| + |\mathcal{S}_{\text{coarse}}^{(t)}|},
}
\label{eq:adaptive_beta}
\end{equation}
where $\beta_{\max} = 0.25$ caps the maximum weight. This implements a natural curriculum—mastering correctness before refining style—without manual schedule design. While curriculum learning has been applied to code generation~\citep{Nair2024curriculum}, our approach uniquely derives the curriculum from execution feedback rather than predefined difficulty metrics.

Following the theoretical analysis in SPIN~\citep{Chen2024spin}, 
our execution-aware filtering effectively redefines $p_{\text{data}}$ 
as the distribution over filtered, execution-validated samples. 
By excluding only execution-failing negatives while preserving semantically equivalent alternatives, 
the SPIN convergence guarantee remains intact: 
the global optimum of $\mathcal{L}$ is achieved if and only if 
$p_\theta = p_{\text{data}}$ under this redefined target, 
ensuring that iterative self-play drives the model toward the target distribution. 
In practice, the opponent generates increasingly sophisticated formulas across 
iterations---from simple errors and misuses early on to subtle semantic 
mistakes later---forcing the main player to develop progressively 
finer-grained discrimination capabilities. 
Figure~\ref{fig:iteration_comparison} visualizes this convergence, showing diminishing marginal gains beyond iteration 3.

\subsection{Execution-based Voting}

Unlike best-of-$N$ sampling~\citep{Nakano2021,Cobbe2021}, which requires external verifiers and additional supervision, \textit{ExecVote} leverages the deterministic constraints of formula execution as a training-free validity filter. While self-consistency methods ~\citep{Wang2023selfconsistency} count surface-level matches, we group candidates based on their semantic behavior during execution. By doing so, we capture the collective strength of semantically equivalent programs~\citep{Chen2021codex} that would otherwise be treated as distinct incorrect samples.

Specifically, we sample $K$ candidate formulas at temperature $T > 1$. Non-executable samples are discarded, and voting is conducted on the resulting execution outputs. We select the highest-probability formula from the majority result set; if no consensus exists, the most probable executable formula is returned, falling back to the overall top-1 candidate if all executions fail.

\begin{algorithm}[h]
\caption{FormulaSPIN Training}
\label{alg:training}
\begin{algorithmic}[1]
\Require Dataset $D$, model $\theta_0$, iterations $K$
\Ensure Trained model $\theta_K$
\For{$t = 0$ to $K-1$}
    \State $S \gets \emptyset$
    \For{each $(q, T, f) \in D$}
        \State $f' \sim p_{\theta_t}(\cdot \mid q, T)$
        \State $r, r' \gets \mathcal{E}(f, T), \mathcal{E}(f', T)$
        \State \textbf{skip if} $r'{=}\texttt{ERR}$ or $f'{=}f$
        \State $c \gets$ \textsc{Fine} if $r'{=}r$ else \textsc{Coarse}
        \State $S \gets S \cup \{(q, T, f, f', c)\}$
    \EndFor
    \State $n_f \gets |\{s \in S: c{=}\textsc{Fine}\}|$
    \State $w_m \gets \beta_{\max} n_f / |S|$
    \For{each batch $B \subseteq S$}
        \State Compute $L$ via Eq.~(1)
        \State Update $\theta$ by gradient descent
    \EndFor
    \State $\theta_{t+1} \gets \theta$
\EndFor
\State \Return $\theta_K$
\end{algorithmic}
\end{algorithm}

\section{Experiments}

\subsection{Experimental Setup}

\textbf{Benchmarks.} We evaluate on two benchmarks: NL2Formula-70K \citep{Zhao2024nl2formula}, which contains 70,799 query–formula pairs from 21,670 tables covering 37 function types and uses the official train/validation/test split of 75\%/10\%/15\%, and Sheetpedia-Selected\citep{Tian2025sheetpedia}, which consists of 2,167 high-quality, human-verified examples drawn from a 290K spreadsheet corpus. A detailed statistics analysis can be found in Appendix~\ref{sec:dataset_stats}.

\vspace{6pt}
\noindent
\textbf{Training Configuration.}
All models are initialized from LLaMA-3.1-8B-Instruct \citep{Dubey2024llama3}. We first obtain an SFT base model on the NL2Formula training set using LoRA ($r=16$, $\alpha=16$, dropout $=0$) and AdamW-8bit, with learning rate $3\times10^{-4}$, weight decay $10^{-3}$, cosine scheduling, and warmup ratio $0.03$. Starting from this SFT-initialized adapter, we perform SPIN self-play fine-tuning for four iterations. In multiple iterations, we leverage the synthetic data from the most recent iteration and add to the newly generated synthetic data, therefore resulting in a synthetic dataset size of 50k at iteration 0 and 100k at iteration 1, 2 and 3. In each SPIN iteration, we train for 2 epochs with per-device batch size 1, gradient accumulation 16, BF16, RMSProp, zero weight decay, and warmup ratio $0.1$. The peak learning rate is $5\times10^{-7}$ for iterations 0--2 and $1\times10^{-7}$ for iteration 3. The SPIN logit scaling coefficient is set to $0.1$ for iterations 0--2 and $5.0$ for the final iteration. Formula execution is conducted using a Linux-based customized spreadsheet formula execution simulator.

\vspace{6pt} 
\noindent
\textbf{Baselines.} We compare against several representative approaches, including FORTAP \citep{Cheng2022fortap}, which builds on TUTA \citep{Wang2021tuta} and extends table pre-training to include spreadsheet formulas, using a two-stage LSTM decoder \citep{Hochreiter1997}; fCODER-Base and fCODER-Large \citep{Zhao2024nl2formula}, which are T5-based sequence-to-sequence models trained with supervised fine-tuning; SFT, our LLaMA-3.1-8B-Instruct \citep{Dubey2024llama3} base model trained with standard supervised fine-tuning for two epochs; SFT (Continuous), which continues training LLaMA-3.1-8B-Instruct for an additional three epochs on the same data; SFT-DPO, which further trains the SFT model using Direct Preference Optimization \citep{Rafailov2023dpo} with wrong–correct pairs as preference data; and a suite of strong proprietary API models evaluated in a zero-shot setting, including GPT-4o \citep{OpenAI2024gpt4o}, Claude-3.7-Sonnet \citep{Anthropic2025claude37}, Gemini-2.5-Pro \citep{Google2025gemini25}, and DeepSeek-R1 \citep{DeepSeek2025r1}.

\vspace{6pt}
\noindent
\textbf{Evaluation Metrics.} We evaluate model performance using four complementary metrics: Exact Match (EM), string-level exact correctness of formulas; Execution Accuracy (EA), semantic correctness via identical execution results; 
Execution Success Rate (ESR), percentage of executable formulas; 
and Formula Sketch Match (FSM), function-level correctness ignoring cell references. 
Following \citet{Zhao2024nl2formula}, we also report results by formula complexity: 
Simple (1–2 functions), Medium (3–4), and Complex (5+). 
Note that this complexity-based categorization is distinct from the Trivial/Fine/Coarse error granularity used during training, which reflects semantic rather than structural distinctions.

\subsection{Main Results}

\textbf{Overall Performance.} Table~\ref{tab:main_results} shows results on the NL2Formula-70K test set. FormulaSPIN achieves 74.9\% EM and 87.1\% EA, substantially outperforming all baselines. All reported results are averaged over three runs with different random seeds. We highlight several key findings:

\begin{table}[t]
\centering
\small
\begin{tabular}{lcccc}
\toprule
\textbf{Model} & \textbf{EM} & \textbf{EA} & \textbf{ESR} & \textbf{FSM} \\
\midrule
FORTAP  & 24.2 & - & - & 58.4 \\
fCODER-Large & 70.6 & 77.1 & 96.5 & 95.1 \\
SFT (base) & 68.2 & 81.1 & 97.1 & 95.7 \\
SFT (Continued) & 68.8 & 79.2 & \textbf{99.4} & 92.6 \\
SFT-DPO & 72.2 & 84.5 & 98.6 & 94.9 \\
\midrule
\multicolumn{5}{l}{\textit{API Models}} \\
DeepSeek-R1       & 34.2 & 62.0 & 88.6 & - \\
Gemini-2.5-Pro    & 33.0 & 68.2 & 92.4 & - \\
Claude-3.7-Sonnet & 32.4 & 61.2 & 87.1 & - \\
GPT-4o            & 34.0 & 66.1 & 91.3 & - \\
\midrule
\multicolumn{5}{l}{\textit{Ours}} \\
\textbf{FormulaSPIN ($t_{0}$)} & 70.1 & 82.5 & 98.8 & 96.1 \\
\textbf{FormulaSPIN ($t_{1}$)} & 72.5 & 84.7 & 99.1 & 97.8 \\
\textbf{FormulaSPIN ($t_{2}$)} & 74.1 & 86.9 & 99.1 & 98.3 \\
\textbf{FormulaSPIN ($t_{3}$)} & \textbf{74.9} & \textbf{87.1} & 99.2 & \textbf{98.7} \\
\bottomrule
\end{tabular}
\captionsetup{font=small}
\caption[Performance on NL2Formula-70K test set.]{Performance on NL2Formula-70K test set.
\textbf{EM}: Exact Match;
\textbf{EA}: Execution Accuracy;
\textbf{ESR}: Execution Success Rate;
\textbf{FSM}: Formula Sketch Match.}
\label{tab:main_results}
\end{table}

\begin{itemize}[nosep,leftmargin=*]
\item  Self-play provides consistent gains: +1.9\% EM at iter $0$, accumulating to +6.7\% EM by iter $3$ over the SFT base.
\item  FormulaSPIN (iter $3$) outperforms SFT-DPO despite using no additional data. DPO uses GPT-4.1 for preference data, yet still underperforms, suggesting that execution-based intrinsic feedback is superior to distillation from external LLMs.
\item  Continued SFT degrades performance (-1.9\% EA), confirming naive multi-epoch training ineffective.
\item  Even large proprietary models underperform our approach, showing the value of task-specific self-play fine-tuning.
\end{itemize}

\begin{figure}[h]
\centering
\includegraphics[width=0.5\textwidth]{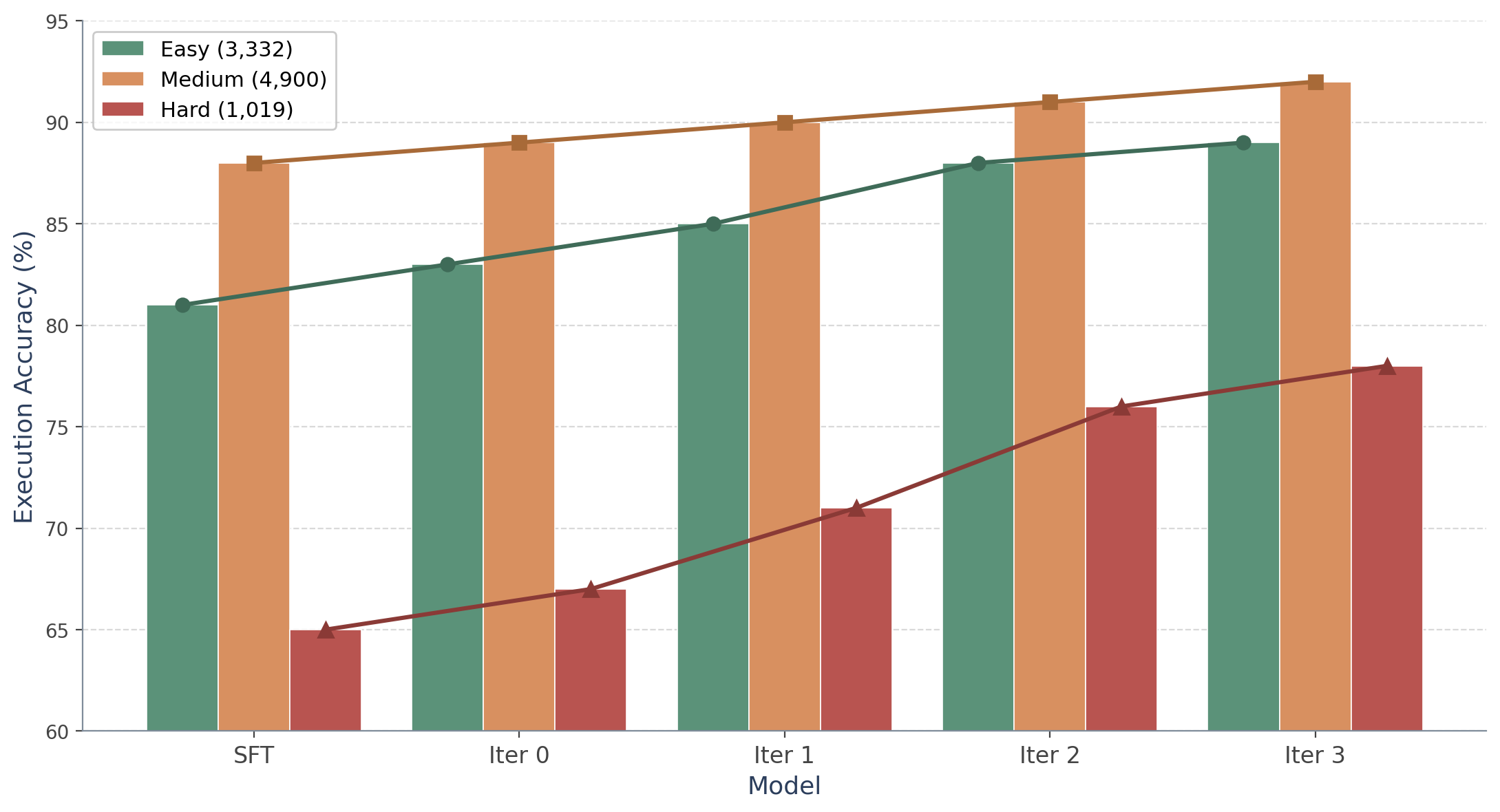}
\caption{Execution Accuracy by formula complexity.}
\label{fig:complexity}
\end{figure}
\vspace{6pt}
\noindent
\textbf{Performance by Complexity.} Figure~\ref{fig:complexity} analyzes results across formula complexity levels. The largest improvements occur on complex formulas (+11.0\% from SFT to iter~$3$), with the bulk of the gain concentrated in iterations~1 and~2 (+4.0 and +5.0 points, respectively) before plateauing at iter~3---suggesting self-play effectively discovers compositional patterns that SFT alone fails to capture. Simple formulas also benefit notably (+8.0\%), while medium formulas, which already attain the highest absolute accuracy under SFT, exhibit the smallest headroom and gain a more modest +4.0\%. This pattern indicates that self-play is most beneficial precisely where SFT struggles the most, narrowing the gap between difficulty tiers as iterations progress.

\vspace{6pt}

\subsection{Generalization and Scaling}
\noindent
\textbf{Execution-based Voting.} Table~\ref{tab:testtime} shows the effect of sampling multiple candidates at inference. Sampling $K=10$ candidates boosts EA by +7.0\% over greedy decoding, at roughly 1.68$\times$ the inference cost. The gains saturate beyond $K=10$, suggesting diminishing returns. This demonstrates that formula generation benefits significantly from test-time compute, similar to code generation \citep{Chen2021codex}.

\begin{table}[H]
\centering
\small
\begin{tabular}{lccc}
\toprule
\textbf{K (samples)} & \textbf{EA} & \textbf{Inference Time} \\
\midrule
1 (greedy) & 86.7 & 1.00$\times$ \\
5          & 91.4 & 1.32$\times$ \\
10          & 93.8 & 1.68$\times$ \\
\bottomrule
\end{tabular}
\caption{Impact of test-time compute scaling (FormulaSPIN iter $3$, 128 test samples). Sampling $K=10$ provides the best accuracy-efficiency tradeoff.}
\label{tab:testtime}
\end{table}

\vspace{6pt}
\noindent
\textbf{Out-of-Domain Generalization.} We evaluate on the Sheetpedia-NL2FL test set (2167 examples from a different data distribution) to assess generalization. The SFT baseline achieves 63.4\% EM and 71.7\% EA, while SFT-DPO improves to 65.3\% EM and 74.2\% EA. FormulaSPIN generalizes well to out-of-domain data, reaching 69.1\% EM and 78.5\% EA. Combined with test-time scaling ($K=10$), performance further increases to 84.9\% EA.

\vspace{6pt}
\noindent
\textbf{Robustness Across Base Models.} To verify that these improvements generalize across model architectures, we also evaluate FormulaSPIN on multiple base models including Qwen3-8B, Qwen2.5-7B, DeepSeek-Coder-7B, and Mistral-7B, observing consistent gains of +6--8\% EM across all architectures.

\subsection{Training Dynamics}
\begin{figure}[h]
\centering
\includegraphics[width=1\columnwidth]{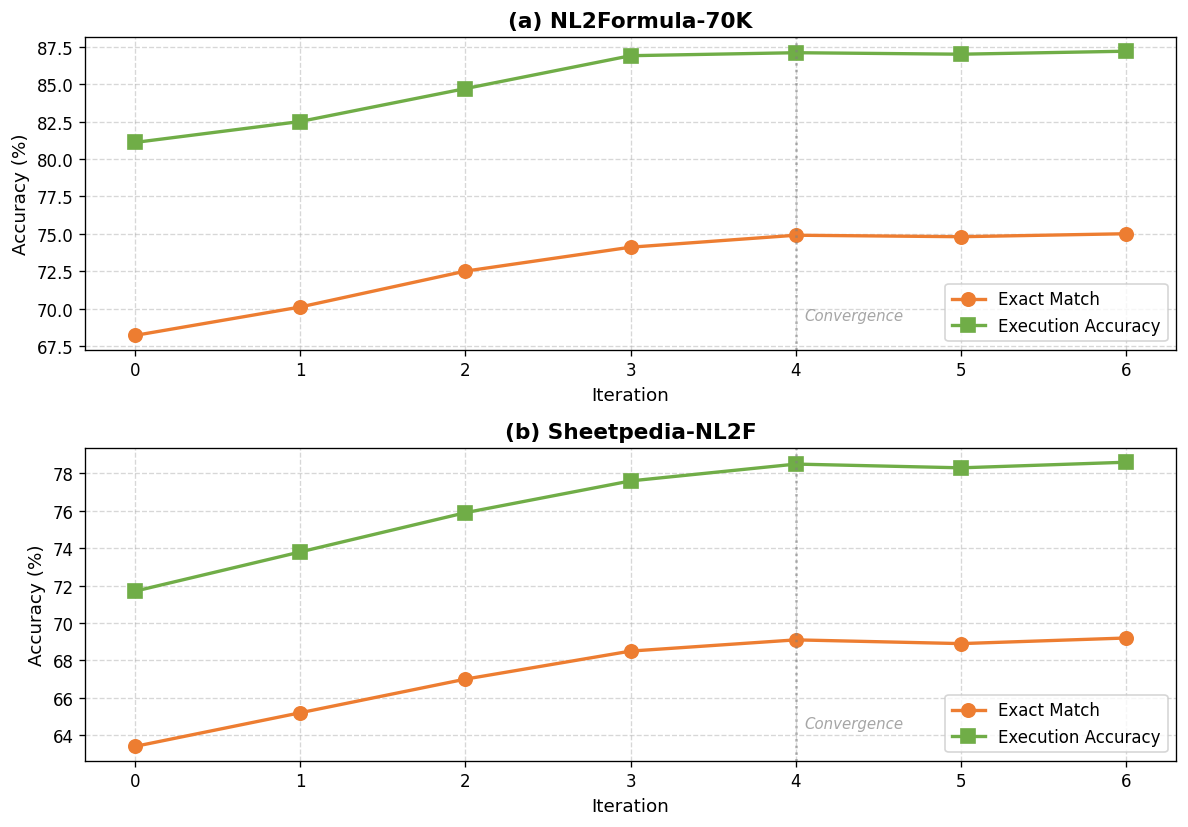}
\caption{Performance progression across iterations on (a) NL2Formula-70K and (b) Sheetpedia-NL2FL. }
\label{fig:iteration_comparison}
\end{figure}

\noindent
\textbf{Iteration Analysis.} Figure~\ref{fig:iteration_comparison} tracks training dynamics across iterations. On NL2Formula, improvements are most pronounced in the early iterations (+2.4\% EM from iter $0$ to 1, +1.6\% from iter 1 to 2) and gradually diminish, with EM rising from 70.1\% at iter 0 to 74.9\% at iter 3 before plateauing. Sheetpedia (out-of-domain) exhibits a similar trajectory, climbing from 63.4\% to 69.1\% EM over the first four iterations and then showing only minor fluctuations around 69\% (69.1\% $\rightarrow$ 68.9\% $\rightarrow$ 69.2\%). Execution Accuracy follows the same pattern on both benchmarks, converging near 87.1\% on NL2Formula and 78.5\% on Sheetpedia. This convergence aligns with theoretical predictions~\citep{Chen2024spin} and indicates that 3--4 iterations provide the optimal trade-off between performance and training cost.

\begin{table}[h]
\centering
\small
\begin{tabular}{lccc}
\toprule
\textbf{Iteration} & \textbf{Exact} & \textbf{Execution} & \textbf{Syntax} \\
& \textbf{Match \%} & \textbf{Match \%} & \textbf{Error \%} \\
\midrule
0 (from SFT) & 71.4 & 83.9 & 4.6 \\
1            & 73.8 & 85.6 & 3.3 \\
2            & 75.2 & 87.4 & 2.5 \\
3            & 76.1 & 88.3 & 1.8 \\
\bottomrule
\end{tabular}
\caption{Quality of self-generated training formulas across iterations. As the model improves, synthetic data quality increases, creating a virtuous cycle.}
\label{tab:synthetic}
\end{table}

\vspace{6pt}
\noindent
\textbf{Synthetic Data Quality.} We analyze the quality of self-generated formulas used in training (Table~\ref{tab:synthetic}). Synthetic data quality improves steadily across iterations: exact matches rise from 71.4\% to 76.1\% (+4.7\%), execution matches climb from 83.9\% to 88.3\% (+4.4\%), and syntax errors drop from 4.6\% to 1.8\%. Notably, synthetic data quality is consistently slightly higher than test-set accuracy as our generation pipeline applies confidence-based filtering and rejection sampling to discard low-quality candidates. This virtuous cycle---a stronger model produces higher-quality training data, which in turn trains an even stronger model---is central to the effectiveness of self-play, and it also explains the diminishing gains observed beyond iteration 3--4: as synthetic data quality approaches the model's own ceiling, the additional learning signal naturally saturates.

\subsection{Ablation Studies}

\textbf{Components Analysis.} Table~\ref{tab:ablation} ablates the key components of FormulaSPIN. Disabling the adaptive curriculum and reverting to a fixed $\beta_{\text{med}}$ leads to a -1.9\% EM and -1.5\% EA degradation, validating that progressively shifting focus from semantic correctness to stylistic refinement is more effective than a static weighting scheme.

\begin{table}[h]
\centering
\small
\begin{tabular}{lcc}
\toprule
\textbf{Configuration} & \textbf{EM} & \textbf{EA} \\
\midrule
FormulaSPIN & \textbf{74.9} & \textbf{87.1} \\
\quad - w/o adaptive curriculum & 73.0 & 85.6 \\
\quad - w/ vanilla SPIN & 70.6 & 82.4 \\
\quad - w/o self-play (SFT only) & 68.2 & 81.1 \\
\bottomrule
\end{tabular}
\caption{Ablation study on NL2Formula-70K removing key components of FormulaSPIN. The adaptive curriculum contributes substantially, while vanilla SPIN degrades performance after the first iteration.}
\label{tab:ablation}
\end{table}

Most strikingly, vanilla SPIN starts \textit{degradeing} performance after iter0. This counterintuitive result arises because vanilla SPIN treats all non-matching outputs uniformly as negatives, penalizing execution-equivalent alternatives (e.g., \texttt{SUM(A1:A5)} vs.\ \texttt{A1+A2+A3+A4+A5}) alongside genuinely incorrect formulas. The resulting contradictory gradients destabilize training and erode the gains achievable through self-play. By contrast, FormulaSPIN's formula-aware self-play objective combined with the adaptive curriculum yields a +6.7\% EM and +6.0\% EA improvement over SFT, demonstrating that both components are indispensable for unlocking the benefits of self-play in formula generation. A more detailed analysis of vanilla SPIN's failure modes is provided in Appendix~\ref{sec:vanilla_spin_analysis}.

\vspace{6pt} 
\noindent
\textbf{Adaptive Curriculum Analysis.} We validate our adaptive weighting mechanism (Equation~\ref{eq:adaptive_beta}) against fixed $\beta_{\text{med}}$ values. The adaptive approach outperforms both fixed weights and a hand-tuned linear schedule. The key advantage is that $\beta_{\text{med}}$ automatically tracks training progress: it starts low  when Coarse samples dominate, then increases as Fine samples become prevalent. This implements a natural semantic-to-stylistic curriculum without manual tuning. 

\begin{table}[h]
\centering
\small
\begin{tabular}{lcc}
\toprule
\textbf{Schedule} & \textbf{EM} & \textbf{EA} \\
\midrule
Fixed $\beta = 0.0$  & 70.1 & 86.9 \\
Fixed $\beta = 0.10$ & 72.3 & 85.4 \\

\textbf{Adaptive} & \textbf{74.9} & \textbf{87.1} \\
\bottomrule
\end{tabular}
\caption{Comparison of $\beta_{\text{med}}$ scheduling strategies.}
\label{tab:curriculum_brief}
\end{table}

\subsection{Qualitative Analysis}

\begin{figure}[h]
\centering
\includegraphics[width=\columnwidth]{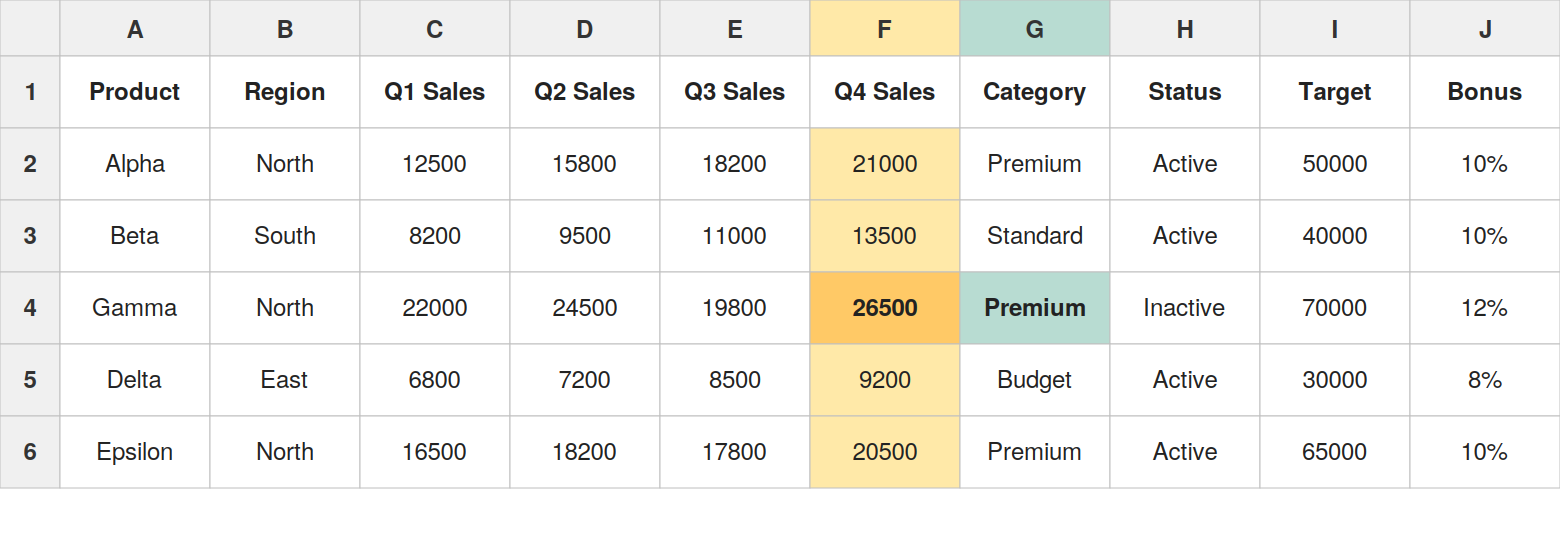}
\caption{An example where SFT produces a legacy array formula requiring special entry mode, while FormulaSPIN generates the modern idiomatic function.}
\label{fig:case_study}
\end{figure}

\vspace{6pt} 
\noindent
\textbf{Case Study.} Figure~\ref{fig:case_study} presents a case study on conditional aggregation. The table in \texttt{A1:J6} contains the NL query ``\textit{What is the highest Q4 sales among Premium category products?}'' The ground-truth formula is \texttt{MAXIFS(F2:F6, G2:G6, "Premium")}, yielding ``26500''. The SFT model predicts \texttt{MAX(IF(G2:G6="Premium", F2:F6))}, a legacy array formula that requires Ctrl+Shift+Enter in older Excel versions and fails silently in certain environments. While FormulaSPIN learns the modern \texttt{MAXIFS} pattern through self-play despite the sparsity of relevant data.

\vspace{6pt}
\noindent
\textbf{Error Analysis.} We conduct stratified random sampling (300 examples each from Simple, Medium, and Complex buckets) on FormulaSPIN ($t_3$) outputs and find that both \emph{error categories} and \emph{function-family involvement} shift systematically with formula difficulty. In the \textit{Simple} bucket, failures concentrate on aggregation choice and over-complication---confusions involving \texttt{SUM} as either reference or prediction account for 29.2\% of errors (e.g., \texttt{SUM}$\rightarrow$\texttt{MAX}, \texttt{SUM}$\rightarrow$\texttt{ROWS}, \texttt{SUM}$\rightarrow$\texttt{AVERAGE}), and the model often rewrites a direct \texttt{FILTER} as a nested \texttt{LET+SUMMARIZE} (\texttt{LET}$\rightarrow$\texttt{FILTER}, 12.5\%). In the \textit{Medium} bucket, the model essentially stops picking the wrong family: errors remain inside \texttt{UNIQUE}$\rightarrow$\texttt{UNIQUE} (35.3\%) and \texttt{SORT}$\rightarrow$\texttt{SORT} (23.5\%) confusions and localize to \emph{projection and value binding}---row selection is typically correct, but the model returns the wrong column (17.6\%) or fails to match surface forms such as score strings, Unicode punctuation, and decimals like \texttt{.9} versus \texttt{0.9} (29.4\%). In the \textit{Complex} bucket, \texttt{LET}$\rightarrow$\texttt{LET} confusions alone account for 63.8\% of errors: the outer program skeleton is largely correct, but inner field ordering and condition bindings are misplaced (\textit{wrong\_return\_column} 27.7\%, \textit{wrong\_filter\_value\_or\_threshold} 17.0\%). This trajectory---from family-level mistakes and template over-complication, to within-family projection drift, to fine-grained binding drift inside correct skeletons---mirrors the semantic-to-stylistic curriculum learned during self-play: the model first masters \emph{which} program to write, then \emph{what} to return, and finally struggles only with \emph{how} bindings are wired inside complex compositions. A detailed breakdown by bucket, error category, function-confusion pattern, and representative case studies appears in Appendix~\ref{sec:error_analysis}.

\section{Conclusion}
In this paper, we propose a novel self-play fine-tuning framework for natural 
language to spreadsheet formula generation, called FormulaSPIN. FormulaSPIN 
leverages the intrinsic verifiability of formulas to enable iterative 
self-improvement without any additional costly human annotations or external 
teacher models. Unlike vanilla SPIN, it avoids contradictory gradients on 
execution-equivalent samples by \textit{distinguishing semantic 
correctness from stylistic variations} through error-aware 
adversarial training. The proposed adaptive curriculum and semantic consensus voting show that task-intrinsic verification can effectively drive 
self-improvement. Our work represents the first effective application of 
self-play to formula generation tasks, significantly outperforming SFT, domain-specific models, and large proprietary models while matching 
models trained with large-scale preference annotations. These findings 
underscore the strong potential of combining self-play with task-intrinsic and dynamic
feedback, opening promising directions for extending this execution-driven 
paradigm to other structured generation tasks with potentially sparse data, including SQL synthesis, code 
generation, and mathematical reasoning.

\section*{Limitations}

Our approach assumes access to a spreadsheet execution engine for validation, which may not be available in all deployment scenarios. The test-time compute scaling improves accuracy but increases inference latency, potentially limiting real-time applications. Additionally, self-play requires multiple training iterations , which may be prohibitive for resource-constrained settings. Finally, our evaluation focuses on English queries and Excel formulas; generalization to other languages and spreadsheet applications remains unexplored.

\section*{Ethics Statement}

This work uses publicly available benchmarks (NL2Formula-70K and Sheetpedia) containing synthetic spreadsheet examples without personally identifiable information. FormulaSPIN is designed to democratize spreadsheet usage by lowering technical barriers for non-expert users. We acknowledge that users should verify generated formulas before deployment in high-stakes scenarios, as over-reliance on automated outputs may introduce errors. Our self-play approach eliminates dependence on expensive proprietary models for preference annotation, reducing both financial and environmental costs compared to methods requiring extensive API calls.

\bibliography{custom}

\appendix

\section{Dataset Statistics}
\label{sec:dataset_stats}
\begin{table}[H]
\centering
\small
\begin{tabular}{lcc}
\toprule
\textbf{Statistic} & \textbf{NL2F-70K} & \textbf{Sheetpedia (OOD)} \\
\midrule
Total examples & 70,799 & 2,167 \\
Unique tables & 21,670 & 1,847 \\
Function types & 37 & 42 \\
Avg.\ query len. & 11.2 & 15.7 \\
Avg.\ formula len. & 10.2 & 15.3 \\
Avg.\ table rows & 10.8 & 14.2 \\
Avg.\ table columns & 6.0 & 9.3 \\
\midrule
\multicolumn{3}{l}{\textit{Complexity Distribution}} \\
\midrule
Simple (1--2 funcs) & \multicolumn{2}{c}{29.1\%} \\
Medium (3--4 funcs) & \multicolumn{2}{c}{58.3\%} \\
Complex (5+ funcs) & \multicolumn{2}{c}{12.6\%} \\
\bottomrule
\end{tabular}
\caption{Dataset statistics for NL2Formula-70K and Sheetpedia-NL2F benchmarks.}
\label{tab:dataset_stats}
\end{table}

\textbf{Dataset Differences.} NL2Formula-70K and Sheetpedia differ fundamentally in their construction and content characteristics. NL2F-70K is derived by converting text-to-SQL examples (WikiSQL, Spider) into spreadsheet contexts, which limits its coverage to SQL-compatible patterns. In contrast, Sheetpedia is curated from real-world sources including enterprise email archives (Enron) and user-contributed content from ExcelForum, capturing authentic spreadsheet usage across diverse domains such as financial modeling and data analysis. Consequently, Sheetpedia exhibits native Excel formula patterns with prevalent error-handling constructs (e.g., IFERROR wrapping VLOOKUP), nested conditional logic, and complex function co-occurrences that are rarely observed in SQL-converted datasets. This makes Sheetpedia a challenging out-of-distribution (OOD) benchmark for evaluating model generalization to real-world spreadsheet scenarios.

\section{DPO Baseline Construction}
\label{sec:dpo_baseline}

\paragraph{Dataset Quality Analysis.}
The NL2Formula-70K dataset \citep{Zhao2024nl2formula} was collected from web sources, where the ground-truth formulas represent one possible solution rather than the canonical or most efficient implementation. Through careful manual inspection of 500 randomly sampled examples, we observed that approximately 18.3\% of ground-truth formulas could be expressed more concisely or idiomatically.

For instance, consider the following examples where SFT-generated formulas are arguably superior:

\begin{table}[H]
\centering
\small
\begin{tabular}{p{2.8cm}p{4.2cm}}
\toprule
\textbf{Ground Truth} & \textbf{SFT Output (More Efficient)} \\
\midrule
\texttt{A1+A2+A3+A4+A5} & \texttt{SUM(A1:A5)} \\
\midrule
\texttt{SUMPRODUCT((B:B} \newline \texttt{="X")*(C:C))} & \texttt{SUMIF(B:B,"X",C:C)} \\
\midrule
\texttt{INDEX(A:A, MATCH(MAX(B:B), B:B,0))} & \texttt{XLOOKUP(MAX(B:B), B:B,A:A)} \\
\bottomrule
\end{tabular}
\caption{Examples where alternative formulas may be more efficient or idiomatic than ground-truth annotations.}
\label{tab:formula_alternatives}
\end{table}

\paragraph{Motivation for Two-Stage Annotation.}
Given this observation, naively setting the ground-truth formula as always preferred in DPO training could lead to suboptimal learning dynamics. The model might learn to memorize specific formula patterns from the dataset rather than developing genuine preferences for efficient, correct formulas.

\paragraph{Two-Stage Preference Annotation Strategy.}
To construct a stronger and more principled DPO baseline, we adopt a two-stage preference annotation strategy:

\textbf{Stage 1: Execution Validation.} For each training example $(q, \mathcal{T}, f_{gt})$, we first generate a candidate formula $f_{sft}$ using the SFT model. Both $f_{gt}$ and $f_{sft}$ are executed on the table $\mathcal{T}$. If $\mathcal{E}(f_{sft}, \mathcal{T}) \neq \mathcal{E}(f_{gt}, \mathcal{T})$ (execution results differ), we set the ground-truth as preferred: $(f_w, f_l) = (f_{gt}, f_{sft})$. If $\mathcal{E}(f_{sft}, \mathcal{T}) = \mathcal{E}(f_{gt}, \mathcal{T})$ (execution results match), we proceed to Stage 2.

\textbf{Stage 2: GPT-4.1 Preference Labeling.} When both formulas produce identical execution results, we employ GPT-4.1 to determine which formula is preferred based on criteria including conciseness, readability, idiomatic usage, and computational efficiency. 

\paragraph{Strategy Benefits.}
This strategy ensures that semantically incorrect formulas are never favored over correct ones, that among multiple valid solutions the more efficient or idiomatic variant is preferred, and that the DPO baseline learns genuinely meaningful preferences rather than relying on arbitrary memorization.

\paragraph{Resulting Dataset Statistics.}
The resulting preference pairs consist of approximately 71.2\% cases where ground-truth was preferred (due to execution mismatch or GPT-4.1 preference) and 28.8\% cases where the SFT-generated formula was deemed superior.

\section{Why Vanilla SPIN Fails for Formula Generation}
\label{sec:vanilla_spin_analysis}

We provide detailed analysis explaining why vanilla SPIN~\citep{Chen2024spin} is unsuitable for formula generation, resulting in performance degradation compared to FormulaSPIN.

\paragraph{Iteration-wise Performance.}
Table~\ref{tab:vanilla_iterations} tracks vanilla SPIN across iterations. Unlike FormulaSPIN which shows consistent improvement, vanilla SPIN peaks at iteration 1 with marginal gains, then \textit{degrades} in subsequent iterations.

\begin{table}[H]
\centering
\small
\begin{tabular}{lcccc}
\toprule
\textbf{Method} & \textbf{Iter 0} & \textbf{Iter 1} & \textbf{Iter 2} & \textbf{Iter 3} \\
\midrule
Vanilla SPIN & 68.5 & 71.9 & 71.1 & 70.6 \\
FormulaSPIN  & 68.5 & 72.5 & 74.1 & 74.9  \\
\bottomrule
\end{tabular}
\caption{Exact Match (\%) across iterations on NL2Formula-70K. Vanilla SPIN shows early saturation and subsequent degradation, while FormulaSPIN improves consistently.}
\label{tab:vanilla_iterations}
\end{table}

\paragraph{Core Issues with Vanilla SPIN.}
We identify three fundamental problems that cause vanilla SPIN to fail for formula generation tasks.

\begin{itemize}
    \item {Problem 1: Penalizing Execution-Equivalent Formulas.}
    Vanilla SPIN treats all $f' \neq f$ as negative samples. However, in formula generation, many alternatives are semantically correct.In our training data, approximately 15-20\% of generated formulas at each iteration are execution-equivalent to references. Vanilla SPIN incorrectly pushes the model \textit{away} from these valid solutions, creating contradictory learning signals.

    \item {Problem 2: Noise from Syntax Errors.}
    Vanilla SPIN includes all generated samples regardless of executability. Early iterations produce 10-12\% syntax errors with highly diverse patterns (missing parentheses, invalid function names, malformed references). These samples provide inconsistent gradients that destabilize training rather than informing it.

    \item {Problem 3: Conflating Semantics and Style.}
    Without distinguishing Coarse (wrong result) from Fine (correct result, different form) samples, vanilla SPIN simultaneously optimizes for: (1) Semantic correctness: preferring formulas that compute the right value; (2) Stylistic conformity: preferring formulas that match reference syntax. These objectives can conflict—a verbose but correct formula should not receive the same penalty as an incorrect one. The uniformity prevents the model from establishing a clear learning hierarchy.
\end{itemize}

\begin{table}[h]
\centering
\small
\begin{tabular}{p{3.2cm}p{3.8cm}}
\toprule
\textbf{Reference} & \textbf{Generated (Equivalent)} \\
\midrule
\texttt{SUM(A1:A5)} & \texttt{A1+A2+A3+A4+A5} \\
\texttt{AVERAGE(B:B)} & \texttt{SUM(B:B)/COUNT(B:B)} \\
\texttt{IF(A1>0,A1,0)} & \texttt{MAX(A1,0)} \\
\bottomrule
\end{tabular}
\caption{Examples of execution-equivalent formula pairs incorrectly penalized by vanilla SPIN.}
\label{tab:equivalent_examples}
\end{table}

\section{Generalization Across Base Models}
\label{sec:model_generalization}
We evaluate FormulaSPIN across multiple base models to demonstrate its generalizability. Table~\ref{tab:full_model_results} presents complete results.
\begin{table}[h]
\centering
\small
\begin{tabular}{lcccc}
\toprule
\textbf{Base Model} & \textbf{Params} & \textbf{SFT} & \textbf{SPIN} & \textbf{$\Delta$} \\
\midrule
LLaMA-3.1-8B & 8B & 68.2 & 74.9 & +6.7 \\
Qwen3-8B & 8B & 69.5 & 76.3 & +6.8 \\
Qwen2.5-7B & 7B & 68.9 & 73.9 & +5.0 \\
DeepSeek-Coder-7B & 7B & 67.4 & 75.8 & +8.4 \\
Mistral-7B-v0.3 & 7B & 66.1 & 72.3 & +6.2 \\
\bottomrule
\end{tabular}
\caption{FormulaSPIN performance (EM \%) across different base models on NL2Formula-70K. Gains are consistent across architectures.}
\label{tab:full_model_results}
\end{table}

FormulaSPIN provides consistent improvements across all model families, confirming the robustness reported in Section~4.3. Code-specialized models (DeepSeek-Coder) and recent instruction-tuned models (Qwen3) tend to perform well, validating the connection between code and formula generation.

\section{Detailed Error Analysis}
\label{sec:error_analysis}

We provide the full methodology, statistics, and qualitative case studies behind the error analysis summarized in the main text. All numbers in this section come from the same 900-example stratified random sample (300 each from Simple, Medium, and Complex buckets) drawn from FormulaSPIN ($t_3$) greedy outputs. The Calculation bucket is excluded as it is dominated by direct numeric reduction and does not interact with the compositional patterns we study here.

\subsection{Sampling and Annotation Methodology}

For each of the three difficulty buckets (Simple: 1--2 functions, Medium: 3--4 functions, Complex: 5+ functions), we uniformly sample 300 examples from the FormulaSPIN ($t_3$) test outputs and manually label every error according to (i) one of the error categories defined in Table~\ref{tab:error_taxonomy}, and (ii) the (reference function $\rightarrow$ predicted function) pair at the top of the formula tree.

\begin{table}[h]
\centering
\small
\setlength{\tabcolsep}{4pt}
\renewcommand{\arraystretch}{1.15}
\begin{tabular}{@{}p{0.34\linewidth}p{0.62\linewidth}@{}}
\toprule
\textbf{Category} & \textbf{Definition} \\
\midrule
\textit{top function mismatch} & The outermost function in the predicted formula differs from the reference (e.g., \texttt{SUM} vs.\ \texttt{ROWS(\allowbreak UNIQUE(\ldots))}). \\
\midrule
\textit{aggregation confusion} & The outer function is a reduction operator but the wrong one (e.g., \texttt{SUM} substituted for \texttt{MAX}, \texttt{AVERAGE}, or \texttt{ROWS}). \\
\midrule
\textit{lookup-strategy confusion} & A query solvable by a single \texttt{FILTER} is rewritten as a nested \texttt{LET+\allowbreak SUMMARIZE} composition, or vice versa. \\
\midrule
\textit{wrong return column} & Row selection and top-level function are correct, but the projected column index inside \texttt{CHOOSECOLS} (or equivalent) is wrong. \\
\midrule
\textit{wrong column binding} & The function selects rows or returns values from the wrong source column entirely (e.g., \texttt{A2:A6} instead of \texttt{B2:B6}). \\
\midrule
\textit{wrong filter value / threshold} & Filter structure is correct but a literal value or comparison threshold differs from the reference, including surface-form mismatches (Unicode, whitespace, decimal formatting). \\
\midrule
\textit{missing or extra filter condition} & The predicate has one fewer or one more conjunct than the reference. \\
\midrule
\textit{structural drift} & The formula tree shape diverges substantially from the reference even though both share top-level intent. \\
\midrule
\textit{same-sketch wrong argument binding} & The function tree matches the reference exactly, but argument order inside a multi-argument call (e.g., \texttt{HSTACK}) is permuted. \\
\bottomrule
\end{tabular}
\caption{Error category taxonomy used to label the 900 stratified samples.}
\label{tab:error_taxonomy}
\end{table}

\subsection{Per-Bucket Statistics}

Table~\ref{tab:bucket_stats} summarizes performance and residual-error rates per bucket. The bulk of remaining headroom lies in the Complex bucket, where execution accuracy drops by roughly 10 points relative to Simple and Medium, while Simple and Medium errors are sparse and increasingly localized.

\begin{table}[H]
\centering
\small
\begin{tabular}{lcccc}
\toprule
\textbf{Bucket} & \textbf{EM} & \textbf{EA} & \textbf{ESR} & \textbf{Errors} \\
\midrule
Simple  & 59.0 & 92.0 & 100.0 & 24 (8.0\%)  \\
Medium  & 82.0 & 94.3 & 100.0 & 17 (5.7\%)  \\
Complex & 82.7 & 84.3 & 97.0  & 47 (15.7\%) \\
\bottomrule
\end{tabular}
\caption{Per-bucket statistics on the 900-example stratified sample.}
\label{tab:bucket_stats}
\end{table}

\subsection{Error Category Distribution}

We group errors into the categories defined above. Table~\ref{tab:error_distribution_full} reports their distribution per bucket; bold numbers highlight the dominant category in each column. The shape of the distribution changes qualitatively across buckets: Simple errors are spread across template-selection categories (\textit{top function mismatch}, \textit{aggregation confusion}, \textit{lookup-strategy confusion}); Medium errors collapse onto \textit{top function mismatch} and surface-form mismatches; Complex errors localize to \textit{wrong return column} and \textit{wrong filter value / threshold} inside otherwise correct \texttt{LET} skeletons.

\begin{table}[H]
\centering
\small
\setlength{\tabcolsep}{4pt}
\begin{tabular}{@{}lccc@{}}
\toprule
\textbf{Error category} & \textbf{Simple} & \textbf{Medium} & \textbf{Complex} \\
\midrule
top function mismatch          & \textbf{29.2} & \textbf{41.2} & 23.4 \\
aggregation confusion          & 16.7 & --   & --   \\
lookup-strategy confusion      & 16.7 & --   & --   \\
wrong column binding           & 12.5 & 11.8 & 4.3  \\
wrong filter value             & 8.3  & 29.4 & 17.0 \\
wrong return column            & --   & 17.6 & \textbf{27.7} \\
missing/extra filter cond.     & 8.3  & --   & 4.3  \\
structural drift               & 4.2  & --   & --   \\
same-sketch arg.\ binding      & 4.2  & --   & --   \\
\bottomrule
\end{tabular}
\caption{Error category distribution (\% of bucket errors) for FormulaSPIN ($t_3$). ``--'' indicates the category did not appear in that bucket's sample.}
\label{tab:error_distribution_full}
\end{table}

\subsection{Function Confusion Patterns}

Table~\ref{tab:func_confusion_full} reports the top (reference $\rightarrow$ predicted) function pairs per bucket. The Simple bucket is dominated by \texttt{SUM}-related confusions (29.2\%) and \texttt{LET}$\rightarrow$\texttt{FILTER} over-complication (12.5\%). The Medium bucket is dominated by within-family confusions: \texttt{UNIQUE}$\rightarrow$\texttt{UNIQUE} (35.3\%) and \texttt{SORT}$\rightarrow$\texttt{SORT} (23.5\%), meaning the model picks the right family but mis-projects. The Complex bucket is dominated by \texttt{LET}$\rightarrow$\texttt{LET} (63.8\%), meaning the model picks the right top-level template but misplaces inner bindings.

\begin{table}[h]
\centering
\small
\begin{tabular}{lc}
\toprule
\textbf{Top confusion (ref $\rightarrow$ pred)} & \textbf{\% of bucket errors} \\
\midrule
\multicolumn{2}{l}{\textit{Simple}} \\
\texttt{SUM}-family confusions (combined)   & 29.2 \\
\texttt{LET}$\rightarrow$\texttt{FILTER}    & 12.5 \\
\texttt{SUMIFS}$\rightarrow$\texttt{ROWS}   & 8.3  \\
\midrule
\multicolumn{2}{l}{\textit{Medium}} \\
\texttt{UNIQUE}$\rightarrow$\texttt{UNIQUE} & 35.3 \\
\texttt{SORT}$\rightarrow$\texttt{SORT}     & 23.5 \\
\texttt{SUMIFS}$\rightarrow$\texttt{ROWS}   & 17.6 \\
\midrule
\multicolumn{2}{l}{\textit{Complex}} \\
\texttt{LET}$\rightarrow$\texttt{LET}       & 63.8 \\
\texttt{SORT}$\rightarrow$\texttt{SORT}     & 6.4  \\
\texttt{LET}$\rightarrow$\texttt{ROWS}      & 6.4  \\
\bottomrule
\end{tabular}
\caption{Top function-confusion patterns by bucket. The transition from cross-family to within-family to within-template confusions is monotone with difficulty.}
\label{tab:func_confusion_full}
\end{table}

\subsection{Representative Case Studies}

We illustrate each bucket's characteristic failure mode with a representative case drawn from our annotated set.

\paragraph{Simple --- aggregation confusion.}
The model substitutes \texttt{SUM} for the correct reduction operator. Both predictions execute successfully but produce semantically wrong values.
\begin{quote}
\textit{Query:} ``What is the best top 10 when there are fewer than 0 wins?'' \\
\textit{Pred:} \texttt{SUM(FILTER(E2:E12, D2:D12<0))} \\
\textit{Ref:}  \texttt{MAX(FILTER(E1, D1<0))}
\end{quote}

\paragraph{Simple --- over-complication.}
A query solvable by a single \texttt{FILTER} is rewritten as a two-level \texttt{LET+SUMMARIZE} composition.
\begin{quote}
\textit{Query:} ``What is the Duration for less than 53 consecutive wins?'' \\
\textit{Pred:} \texttt{LET(query1, SUMMARIZE(D2:D8, SUMX(B2:B8)),} \\
\hspace*{2.6em}\texttt{FILTER(query1, CHOOSECOLS(query1,2)<53))} \\
\textit{Ref:}  \texttt{FILTER(D1, B1<53)}
\end{quote}

\paragraph{Medium --- surface-form mismatch.}
Row selection is correct, but the literal value mismatches the table due to Unicode/whitespace differences.
\begin{quote}
\textit{Query:} ``\ldots best fit (WMAP only) is .9 $\pm$ .1 \ldots'' \\
\textit{Pred:} filter on \texttt{C2:C11="0.9 $\pm$ 0.1"} \\
\textit{Ref:}  filter on \texttt{C1=".9 $\pm$ .1"}
\end{quote}

\paragraph{Medium --- wrong return column.}
The model selects the right rows and the right top-level function (\texttt{UNIQUE}$\rightarrow$\texttt{UNIQUE}) but projects the wrong column.
\begin{quote}
\textit{Query:} ``Who won the match when the winner used the Pedigree attack?'' \\
\textit{Pred:} \texttt{\ldots CHOOSECOLS(FILTER(\ldots), 2)} \\
\textit{Ref:}  \texttt{\ldots CHOOSECOLS(FILTER(\ldots), 4)}
\end{quote}

\paragraph{Complex --- inner field-ordering inside \texttt{LET}$\rightarrow$\texttt{LET}.}
The outer \texttt{LET} skeleton matches the reference; the \texttt{HSTACK} argument order inside is swapped.
\begin{quote}
\textit{Query:} ``Show ids for all aircrafts with more than 1000 distance.'' \\
\textit{Pred:} \texttt{LET(\ldots, SUMMARIZE(HSTACK(D2:D17, A2:A17), \ldots), \ldots)} \\
\textit{Ref:}  \texttt{LET(\ldots, SUMMARIZE(HSTACK(A2:A17, D2:D17), \ldots), \ldots)}
\end{quote}

\section{Comparison with SPFT-SQL}
\label{sec:comparison_spftsql}

We compare FormulaSPIN with SPFT-SQL \citep{Zhang2025spftsql}, a concurrent work applying self-play to Text-to-SQL. Although both methods adapt SPIN to executable generation tasks, they address fundamentally different problems arising from distinct domain properties.

\paragraph{The Execution-Equivalence Problem is Domain-Specific.}
The core challenge in formula generation is that 15--20\% of model-generated formulas are syntactically different from the ground truth yet produce identical execution results (e.g., \texttt{SUM(A1:A5)} vs.\ \texttt{A1+A2+A3+A4+A5}). Vanilla SPIN penalizes all non-matching outputs uniformly, causing the same canonical form to be rewarded in one example while being pushed away as a negative in another---creating systematic gradient conflicts during training. This failure mode has no analog in Text-to-SQL: syntactically distinct SQL queries rarely produce identical outputs across all database rows, so execution-equivalence is not a prevalent source of training instability.

\paragraph{Solution Granularity.}
FormulaSPIN addresses gradient-level conflicts by rethinking the training objective itself---introducing execution-aware sample categorization (Trivial/Coarse/Fine) with adaptive weighting to prevent contradictory gradients on valid alternatives. SPFT-SQL uses execution feedback as a static data-quality filter and reward signal, but does not confront gradient conflicts during optimization because such conflicts are rare in their domain. Their primary contribution is iterative data synthesis (VBI-FT) to combat overfitting from limited supervision---a complementary concern orthogonal to ours.

\end{document}